\documentclass[a4paper,fleqn]{cas-dc}

\usepackage[authoryear]{natbib}

\usepackage{multirow}%
\usepackage{amsmath,amssymb,amsfonts}
\usepackage{mathrsfs}%
\usepackage{listings}%
\usepackage{graphicx}
\usepackage{pifont}
\usepackage{booktabs}
\usepackage{algorithm,algorithmic}
\usepackage{pifont}
\usepackage{tabularx}
\usepackage{hyperref}

\definecolor{mygray}{RGB}{239, 239, 239}

\def\tsc#1{\csdef{#1}{\textsc{\lowercase{#1}}\xspace}}
\tsc{WGM}
\tsc{QE}

\begin{document}
\let\WriteBookmarks\relax
\def\floatpagepagefraction{1}
\def\textpagefraction{.001}

\shorttitle{}    

\shortauthors{}  

\title [mode = title]{ReactDiff: Latent Diffusion for Facial Reaction Generation}  



%

\author[1]{Jiaming Li}


\fnmark[1]




\affiliation[1]{organization={School of Biomedical Engineering \& State Key Laboratory of Advanced Medical Materials and Devices, ShanghaiTech University},
            city={Shanghai},
            postcode={201210}, 
            state={Shanghai},
            country={China}}
\affiliation[2]{organization={School of Biomedical Engineering, Shanghai Jiao Tong University},
            city={Shanghai},
            postcode={200030}, 
            state={Shanghai},
            country={China}}
\affiliation[3]{organization={Shanghai Clinical Research and Trial Center},
            city={Shanghai},
            postcode={201210}, 
            state={Shanghai},
            country={China}}

\author[1]{Sheng Wang}

\fnmark[1]

\author[2]{Xin Wang}

\author[1]{Yitao Zhu}

\author[1]{Honglin Xiong}

\author[2]{Zixu Zhuang}

\author[1,3]{Qian Wang}[orcid=0000-0002-3490-3836]
\cormark[1]
\ead{qianwang@shanghaitech.edu.cn}




\cortext[1]{Corresponding author}

\fntext[1]{These authors contributed equally to this work.}


\begin{abstract}
Given the audio-visual clip of the speaker, facial reaction generation aims to predict the listener's facial reactions. The challenge lies in capturing the relevance between video and audio while balancing appropriateness, realism, and diversity. While prior works have mostly focused on uni-modal inputs or simplified reaction mappings, recent approaches such as PerFRDiff have explored multi-modal inputs and the one-to-many nature of appropriate reaction mappings. In this work, we propose the Facial Reaction Diffusion (ReactDiff) framework that uniquely integrates a Multi-Modality Transformer with conditional diffusion in the latent space for enhanced reaction generation. Unlike existing methods, ReactDiff leverages intra- and inter-class attention for fine-grained multi-modal interaction, while the latent diffusion process between the encoder and decoder enables diverse yet contextually appropriate outputs. Experimental results demonstrate that ReactDiff significantly outperforms existing approaches, achieving a facial reaction correlation of 0.26 and diversity score of 0.094 while maintaining competitive realism. The code is open-sourced at \href{https://github.com/Hunan-Tiger/ReactDiff}{github}.
\end{abstract}




\maketitle


\section{Introduction}
\label{sec1}
Human-Computer Interaction (HCI) has evolved significantly since its emergence in the 1980s, expanding from simple interface designs to sophisticated emotional and behavioral interactions~\citep{dosovitskiy2020image}.
A critical aspect of modern HCI is the ability to generate appropriate facial reactions in response to human behaviors, which is essential for creating natural and engaging human-agent interactions.
This capability has become increasingly important with the rise of virtual agents in applications ranging from healthcare consulting to customer service, where emotional engagement plays a crucial role in user experience.

\begin{figure}
\centering
\includegraphics[width=1\linewidth]{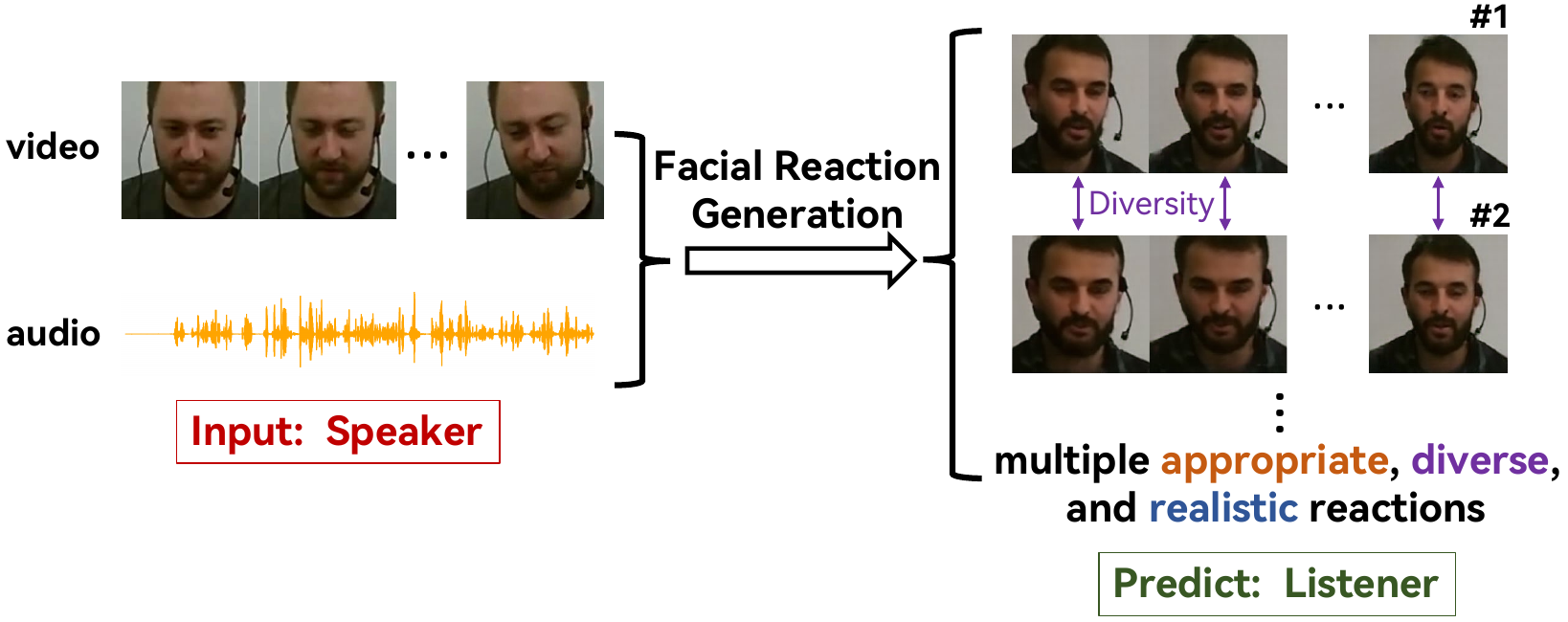}
\caption{Illustration of the facial reaction generation task. Given the speaker's behaviors (video and audio), the goal is to generate multiple listener reactions, ensuring that they are appropriate, realistic, and diverse.}
\label{fig:Facial reaction generation task}
\end{figure}

The development of facial reaction generation has paralleled advances in deep learning.
Early approaches using Convolutional Neural Networks (CNNs) established foundations in facial recognition and generation, such as Generative Adversarial Networks~\citep{goodfellow2014generative}, Variational Autoencoders~\citep{kingma2013auto}, and flow-based models~\citep{dinh2016density}.
The advent of Transformers~\citep{vaswani2017attention} enabled better processing of temporal sequences and multi-modal data, while diffusion models~\citep{song2020denoising, rombach2022high} advanced the quality and variety of image generation.

However, generating appropriate facial reactions remains a significant challenge due to several interrelated factors.
First, there is the complexity of processing multi-modal temporal information, as both audio and visual inputs must be synchronized and interpreted in real time to ensure the facial expressions are contextually accurate.
Second, maintaining temporal consistency is crucial. The facial expressions must evolve smoothly over time to match the ongoing dialogue or emotional cues without breaking immersion.
Third, the same speaker behavior can trigger different real facial reactions expressed by different listeners in real-world dyadic interaction scenarios. The task is further complicated by the inherent one-to-many nature of appropriate reaction mappings. In other words, for any given context, multiple different facial expressions could be equally valid or fitting. This variability requires careful balancing to ensure that the generated expressions are both realistic and emotionally coherent.

CNN-based approaches can struggle with processing temporal sequences and integrating multi-modal information.
While Transformer-based methods are effective at processing sequential data, they struggle with generating diverse and appropriate responses, particularly in situations that require a one-to-many mapping between speaker cues and listener reactions.

At the beginning, \citep{luo2024reactface, song2019exploiting, yoon2022genea} attempt to address these issues but typically focus on either uni-modal inputs or neglect the diversity of appropriate reactions.
In applications, ignoring non-verbal inputs and repeating behaviors when interacting with users significantly impacts the realism of the agent~\citep{liang2023unifarn}. This lack of diversity in facial reactions can result in a less engaging and natural interaction, further diminishing the agent’s perceived authenticity.
Recent works have sought to improve the generation of multiple appropriate facial reactions by modeling a broader range of behaviors and contextual cues.
For instance, BEAMER~\citep{hoque2023beamer} introduces a Transformer-VAE architecture that leverages a behavioral encoder to process both the speaker's behavior and texture. This encoder then maps this information into a latent space, which is used to generate a range of reactions from an appropriate listener. The framework emphasizes context-aware interaction, ensuring that the generated facial reactions are both contextually and emotionally aligned.
Jun et al. ~\citep{yu2023leveraging} further enhance this by incorporating latent diffusion models. By modifying the traditional diffusion framework, they improve the model's ability to capture and model the complex context of a conversation, enabling the generation of diverse and contextually relevant facial expressions. The inherent stochasticity of the diffusion process provides a natural mechanism for the generation of multiple plausible reactions, which enhances the realism and flexibility of the agent's responses.
In a similar vein, Nguyen et al. ~\citep{nguyen2024vector} introduce a novel framework leveraging vector quantized diffusion models for generating multiple appropriate facial reactions in dyadic interactions.
Liu et al. ~\citep{liu2024one} tackle the challenge of generating a range of appropriate reactions using discrete latent variables. Their method explicitly models the one-to-many mapping problem, ensuring that multiple, contextually appropriate facial reactions can be generated from the same initial input, thereby addressing the issue of overfitting to a single reaction type.
Finally, a recent approach by Zhu et al. ~\citep{zhu2024perfrdiff} introduces a personalized approach to this problem. By incorporating weight-editing mechanisms into the generation process, their model allows for fine-tuning of facial reactions based on personalized data, such as user-specific preferences or emotional cues. This approach offers a more tailored and individualized set of reactions, further improving the agent's capacity to engage with users in a more realistic and empathetic manner.

The REACT2023~\citep{song2023react2023} and REACT2024~\citep{song2024react} challenges formalize these requirements by introducing the Multiple Appropriate Reaction Generation problem, which emphasizes the need to generate various appropriate reactions to the same context while maintaining realism.
Specifically, given a speaker’s dyadic interaction video and audio, the model generates multiple sequences of the listener’s facial reactions, comprising 3D Morphable Face Model coefficients, Facial Action Units, Valence and Arousal, and Facial Expression.

The challenges include processing multi-modal information (video and audio) from both the speaker and listener while balancing the \textbf{appropriateness}, \textbf{diversity}, and \textbf{realism} of the generated reactions, as shown in Fig.~\ref{fig:Facial reaction generation task}.

To address these challenges, we propose ReactDiff, a novel framework that seamlessly integrates a Multi-Modality Transformer with a conditional diffusion model. The primary objective of ReactDiff is to effectively handle complex multi-modal data and generate diverse, yet contextually appropriate and realistic, facial reactions for virtual agents. By combining these two powerful components, ReactDiff is designed to strike a delicate balance between \textbf{appropriateness}, \textbf{diversity}, and \textbf{realism} in facial reaction generation, overcoming the limitations of existing models that may struggle to capture all these aspects simultaneously.

ReactDiff utilizes a Multi-Modality Transformer to process and align diverse input data (e.g., audio, video, and textual features) through cross-attention mechanisms, enhancing the model’s ability to capture the nuanced relationships across modalities. The conditional diffusion model generates a wide range of realistic reactions by leveraging stochasticity, offering the flexibility needed for diverse responses. Together, these components ensure that the generated facial reactions are not only contextually appropriate but also diverse and emotionally coherent, significantly improving the realism of agent-user interactions.

We evaluate ReactDiff on the official dataset of the REACT2024 challenge~\citep{song2024react}, which is constructed from the existing RECOLA~\citep{ringeval2013introducing} and NoXI~\citep{cafaro2017noxi} datasets.
Experiments demonstrate that ReactDiff can generate more realistic, appropriate, and diverse facial reactions.
The main contributions of this work are as follows: \begin{enumerate}
\item A novel end-to-end framework ReactDiff that uniquely integrates multi-modal processing with conditional diffusion models in the latent space. Unlike previous approaches that treat multi-modal fusion and generation separately, our framework enables joint optimization of these components, leading to more coherent and contextually appropriate facial reactions.
\item A Multi-Modality Transformer architecture featuring intra- and inter-class attention mechanisms specifically designed for facial reaction generation. The hierarchical attention structure processes facial and acoustic features separately while enabling cross-modal interaction, allowing the model to capture both fine-grained facial movements and broader emotional context.
\item Extensive experimental validation on the REACT2024 challenge datasets that demonstrates significant improvements over state-of-the-art methods. Through detailed ablation studies and comparative analyses, we show that each component of ReactDiff contributes meaningfully to its performance across appropriateness, diversity, and realism metrics.
\end{enumerate}

\section{Background}
\subsection{Facial Reaction Generation}
The study of facial reaction generation emerged with the development of Human-Computer Interaction in the early 1990s. Takeuchi et al. introduced facial displays as a new modality in computer-human interaction~\citep{takeuchi1993communicative}, while Poggi et al. established fundamental theories of non-verbal communication systems, focusing on gaze and facial expressions~\citep{poggi2000performative}. Early computational approaches were developed by Ochs et al., who computed facial expressions using a Facial Expression of Emotions Blending Module~\citep{ochs2005intelligent}, and Zeng et al., who examined methods for understanding human affective behaviour in real-world settings~\citep{zeng2007survey}.

With the advent of deep learning, more sophisticated approaches for facial reaction generation have been proposed. DyadGan~\citep{huang2017dyadgan} and StlyeFaceV~\citep{qiu2022stylefacev} pioneered the use of generative models for facial expressions in dyadic interactions. More recent frameworks like ReactFace~\citep{luo2024reactface} and UniFaRN~\citep{liang2023unifarn} have further advanced the field by addressing the challenges of balancing appropriateness, realism, and diversity. Chen et al. propose a diverse graph transformers model that incorporates both spatial dual-graphs and temporal hyperbolic-graph to capture expression-related structures from facial videos~\citep{chen2024cdgt}.

A significant breakthrough in facial reaction generation came with the development of 3D Morphable Face Models (3DMM)~\citep{Blanz1999AMM}. 
These statistical models have become powerful tools for modeling facial geometry and appearance by capturing facial variability across populations. Recent studies have bridged the gap between 2D and 3D representations by proposing methods to generate 2D facial image sequences from 3D facial behaviours\citep{wang2022faceverse, ren2021pirenderer}, enabling high-quality facial reaction generation through the integration of 2D facial landmarks and 3D facial geometry.

\subsection{Multi-Modal Fusion}
Transformer, proposed by Vaswani et al. in 2017~\citep{vaswani2017attention}, has revolutionized deep learning. Initially designed for natural language processing, its self-attention mechanism allows for efficient modeling of long-range dependencies. The architecture's success quickly spreads to other domains: Vision Transformer (ViT)\citep{dosovitskiy2020image} adapts for image processing, while models like CLIP\citep{radford2021learning} and GPT\citep{radford2018improving} demonstrate effectiveness in multi-modal tasks and large-scale language modeling. 
More recently, Diffusion Transformers (DiT)~\citep{peebles2023scalable} have shown promise in generation tasks, further expanding the architecture's versatility across various AI applications.
Building upon the success of Transformer architecture, numerous approaches have been developed to handle multi-modal fusion tasks. The BERT~\citep{devlin2018bert} architecture is extended to a multi-modal two-stream model, processing visual and textual inputs in separate interacting streams~\citep{lu2019vilbert}. 
Significant progress has also been made in learning transferable representations across modalities. Radford et al.\citep{radford2021learning} introduced CLIP, learning transferable visual representations from natural language supervision. BLIP\citep{li2022blip} further advances this field by employing a bootstrapping approach for vision-language understanding and generation.

Recent research has explored emotion understanding in multi-modal contexts. Kumar et.al~\citep{kumar2022bert} proposed a BERT-based dual-channel system for text emotion recognition that demonstrates the effectiveness of transformer-based architectures in emotional understanding tasks. To address the challenge of incomplete data, Cheng et al. propose a transformer autoencoder that leverages both intra-modality and inter-modality transformers~\citep{cheng2024novel}.
In a similar vein, Sadok et al. proposed a multimodal dynamical variational autoencoder that effectively fuses audiovisual features through disentangled latent representations~\citep{sadok2024multimodal}.

\subsection{Diffusion Models}
Diffusion Models (DMs) represent a significant advancement in image generation. Diffusion Probabilistic Models (DPMs) \citep{ho2020denoising} consist of a forward process (mapping signal to noise) and a reverse process (mapping noise to signal). They can denoise an arbitrary Gaussian noise map to a clean image sample after successive denoising passes proposed learning a function to predict noise using a UNet. 
To make DPMs more practical, Denoising Diffusion Implicit Models (DDIMs)~\citep{song2020denoising} first propose a method of implicit sampling through DPMs and accelerated the sampling speed. 
To address computational challenges, Rombach et al. applied DMs in the latent space of pre-trained autoencoders~\citep{rombach2022high}.
For face image processing, Diffusion Autoencoders (DiffAE) \citep{preechakul2022diffusion} encode facial images into two-part latent codes, using a DPM as the decoder for stochastic variations. Building on this, Diffusion Video Autoencoder \citep{kim2023diffusion} is proposed for human face video editing, offering superior reconstruction performance and disentanglement of identity features.
Latent Diffusion Models (LDMs) have evolved significantly since their inception. Early LDMs, such as Stable Diffusion 1.5 XL, primarily rely on CNN-based architectures~\citep{rombach2022high}. 
However, recent advancements have seen a shift towards pure Transformer structures, as exemplified by the Diffusion Transformers (DiT) model~\citep{peebles2023scalable}.
Furthermore, LDMs have demonstrated the ability to be controlled by multiple modal conditions, allowing for more diverse and specific generation outcomes.
For instance, techniques like ControlNet~\citep{zhang2023adding} have enabled fine-grained control over the generation process using various types of conditioning information.

Our work builds on these foundations, using Transformer to aggregate video and audio for semantic representation, and adopting conditional DDIMs to manipulate the integrated representation.

\section{Method}
\subsection{Technical Overview and Comparison with Existing Methods}
Traditional approaches to facial reaction generation have followed two main paradigms: transformer-based methods that excel at temporal modeling but struggle with diversity, and diffusion-based methods that enable diverse generation but lack fine-grained control over multi-modal interactions. For example, BEAMER~\citep{hoque2023beamer} uses a pure transformer architecture that achieves good temporal consistency but produces limited variation in reactions. Conversely, PerFRDiff~\citep{zhu2024perfrdiff} employs diffusion models that generate diverse outputs but may not maintain consistent temporal relationships between audio and visual features.
ReactDiff advances beyond these approaches through three key innovations. First, while existing methods typically process audio and visual features independently before late fusion, our Multi-Modality Transformer (MMT) employs a hierarchical attention structure with both intra- and inter-cross attention, enabling more nuanced multi-modal interactions. Second, unlike previous work that applies diffusion models directly to the output space, we incorporate the diffusion process in the latent space between MMT's encoder and decoder, allowing for better balance between diversity and temporal consistency. Third, our conditional DDIM uses the speaker's behavior constraint vector to guide the generation process, ensuring that diverse reactions remain contextually appropriate.
This unique architecture allows ReactDiff to maintain the temporal modeling capabilities of transformers while leveraging the generative power of diffusion models, addressing the limitations of previous approaches.

\begin{figure*}[t]
    \centering
    \includegraphics[width=1\linewidth]{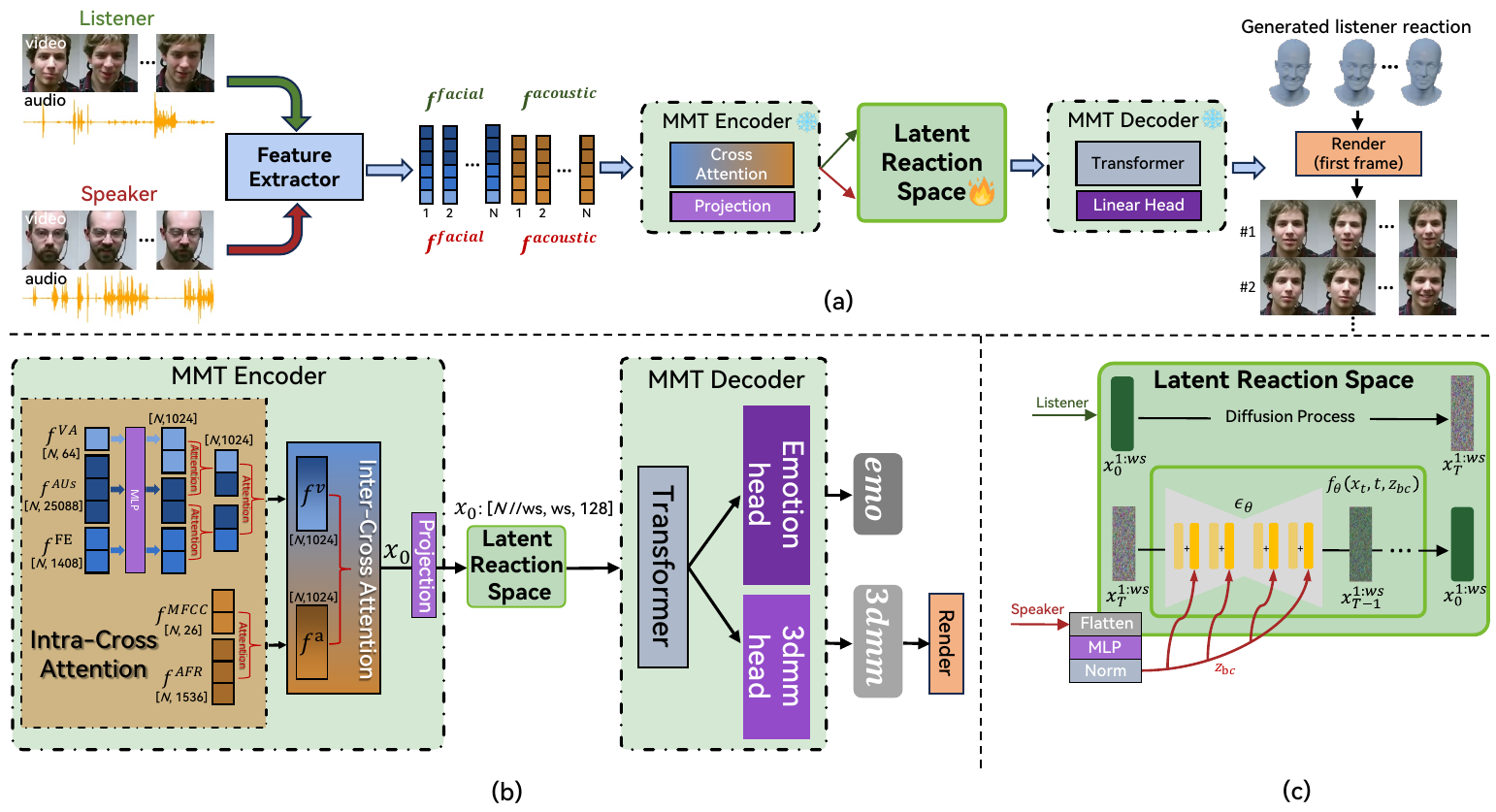}
     \caption{a) Overview of training process of ReactDiff. The extracted video and audio features enter the Latent Reaction Space through the Multi-Modality Transformer, in which the speaker’s behaviour constraint vector $\mathbf{z}_{bc}$ serves as a generation condition.
     b) Overview of MMT. Cross attention not only enhances the internal representation between the modalities but also aligns the semantics between the two modalities, and Transformer models the different window contexts in clip.
     c) Overview of our DDIM. In latent reaction space, crop $x_0$ into several windows, speaker information $z_{bc}$ controls the generation process of listener reactions and finally reshape the denoising token.}
    \label{fig:all}
\end{figure*}

The challenge of generating diverse and realistic facial reactions necessitates advanced modeling techniques. While Variational Autoencoders (VAEs) and Generative Adversarial Networks (GANs) are initially considered, they present limitations in balancing information retention, sampling quality, and training stability. Diffusion Probabilistic Models (DPMs) excel in stability but struggle with multi-modal and temporal information processing~\citep{preechakul2022diffusion}.
To address these limitations, we propose the Facial Reaction Generation Diffusion Model (ReactDiff), integrating cross-attention mechanisms and Transformer layers. ReactDiff comprises three primary modules: feature extraction, Multi-Modality Transformer (MMT), and latent diffusion model.
As illustrated in Fig \ref{fig:all}(a), the framework first extracts facial and acoustic features from raw input videos. The MMT module then reconstructs the listener's reaction, including emotional states and 3D Morphable Model (3DMM) coefficients. Finally, in the latent reaction space, we process the MMT-produced latent vectors to enhance diversity and realism of generated reactions, constrained by corresponding speaker information.
In summary, our design enables complex modeling of relationships within and across modalities. 

\subsection{Feature Extraction}
In ReactDiff, feature extraction is done at every frame, which processes a single image and a 1/25-second audio clip at a time.
Feature extraction of video and audio is done separately. Raw videos are extracted into facial features $f^{facial}$, and audios are extracted into acoustic features $f^{acoustic}$.

\subsubsection{Facial features}
Our ReactDiff framework begins with a crucial step of feature extraction. This process transforms raw input data into meaningful representations that can be effectively utilized by subsequent modules.

For facial reaction generation, raw videos often contain task-irrelevant pixels like background and clothing. 
When trained with limited samples, the model has the risk of being over-fitted (commonly called shortcut learning~\citep{geirhos2020shortcut}) to these irrelevant pixels, ignoring reaction-related features such as facial movements.

First, three facial features are extracted from the video: valence and arousal (denoted as VA), facial action units (denoted as AUs), and facial expression (denoted as FE).
The three variables represent different aspects of facial behaviour and emotions.
\begin{itemize}
\item The VAs are two continuous dimensions used to represent the emotional state.
Valence describes the negativity or positivity of the emotion, ranging from highly negative to highly positive. Arousal represents the intensity or activation level of the emotion, ranging from low (e.g. calm, relaxed) to high (e.g. excited, agitated). 
By controlling the values of valence and arousal, different emotional expressions can be generated, such as happy (e.g. high valence, high arousal) or sad (e.g. low valence, low arousal).
\item The AUs based on the Facial Action Coding System (FACS)~\citep{ekman1978facial} describe the fundamental actions of individuals or groups of muscles on the face. 
Each AU corresponds to a specific muscular activity that produces a distinct facial movement. For example, AU12 represents the contraction of the zygomatic major muscle, which produces a smile.
By controlling the intensity and combination of different AUs, a wide range of facial expressions can be generated.
\item The FEs are categorical labels that describe the overall emotional expression of the face, such as happy, sad, angry, surprised, etc. Each facial expression is typically associated with a specific configuration of AUs and VA values.
\end{itemize}
\par

We use the following pre-trained networks to extract the VA, AU, and FE features of faces in videos.
\begin{itemize}
\item ELIM~\citep{kim2022optimal} is an identity matching method using the Sinkhorn-Knopp algorithm, where reference and target IDs' feature sets are encoded, and optimal matching flows are generated to produce ID-invariant emotional representations. We extract VA features $f^{VA}$ by ELIM encoder. 
\item ME-GraphAU~\citep{luo2022learning} is a method for modeling Action Unit relationships using a combination of node feature learning and multi-dimensional edge feature learning to create an AU relation graph. The AU features $f^{AU}$ are extracted from ME-GraphAU without a classifier.
\item ResMaskNet~\citep{pham2021facial} uses several Masking Blocks, which are applied across Residual Layers to improve the network’s attention ability on important information. The FE features $f^{FE}$ are extracted from ResMaskNet without a classifier. 
\end{itemize}

The dimensions of the extracted features of VA, AU, and FE are 64, 25088, and 1408, respectively.
Ultimately, we get $f^{facial}$ consists of three feature sequences with the shapes of ($N$, 64), ($N$, 25088) and ($N$, 1408), where $N$ means frame number of a clip.

\begin{equation}
\label{eq:Equation 1}
\begin{split}
f^{facial}&=\left \{ f^{VA}, f^{AU}, f^{FE} \right \} .
\end{split}
\end{equation}

\subsubsection{Acoustic features}
We also preprocess the audio for the network. 
Wav2vec 2.0~\citep{baevski2020wav2vec} is a self-supervised framework for learning speech representations directly from raw audio data, achieving high performance on speech recognition tasks with minimal labeled data.
Mel Frequency Cepstral Coefficients (MFCC) features are derived from the Fourier transform of a signal, representing the short-term power spectrum of sound on a mel scale, commonly used in speech and audio processing.
The audio is extracted as MFCC features and Audio Feature Representations (AFR).
The MFCC and AFR are aligned with video frames, as in Equation \ref{eq:Equation 2}.\par
\begin{equation}
\label{eq:Equation 2}
\begin{split}
f^{acoustic}&=\left \{ f^{MFCC}, f^{AFR} \right \} .
\end{split}
\end{equation}
\par
We obtain MFCC and AFR by the TorchAudio and wav2vec 2.0, respectively.
For MFCC, we employ a 16000 Hz sampling rate, 0.04-second window size and stride, and 26 filter banks, yielding 25 frames/sec and a feature dimension of 26. 
For AFR, we use the \textit{wav2vec2-base-960h} version, resulting in a tensor dimension of 1536.

After feature extraction for every frame is finished, we get $f^{acoustic}$ consisting of two feature sequences with the shapes of ($N$, 26) and ($N$, 1536), where $N$ means frame number.

\subsection{Multi-Modality Transformer}
Multi-modal inputs contain rich semantic information, but effectively modeling and utilizing this information has always been a challenge. 
To successfully capture the correlations between input modalities and establish context relations, we propose a highly efficient Multi-Modality Transformer (MMT).
It combines the cross attention mechanism and the modeling capabilities of the Transformer. 
This fusion not only enhances the model's ability to capture complex relationships within and across different modalities but also maintains the temporal consistency of generated facial reactions. In the following, we will delve into the details of MMT.

The MMT adopts an encoder-decoder structure, as shown in Fig. \ref{fig:all}(b). 
To better model sequence contexts and capture dependency between different windows in clip, we use Transformer~\citep{vaswani2017attention} decoder layers to form the backbone of the decoder instead of using CNN layers.
Notably, we also design intra-cross and inter-cross attention layers, respectively, to handle information interaction within and between modalities.

In the encoding phase, the intra-cross attention layers separately process facial features ($f^{facial}$) and acoustic features ($f^{acoustic}$). 
First, The AU features $f^{AU}$ perform intra-cross attention with the FE features $f^{FE}$ and the VA features $f^{VA}$, projecting the results to $f^v\in{\mathbb{R}}^{N\times1024}$. Similarly, $f^{acoustic}$ undergoes processing to obtain $f^a\in{\mathbb{R}}^{N\times1024}$. Next, inter-cross attention is performed between $f^v$ and $f^a$, and projects both onto $F\in{\mathbb{R}}^{N\times128}$. 
Considering the listener's reaction usually depends more on the speaker's reaction at this moment and in the past few seconds, we divide each clip into $n$ windows, with the window size ($ws$) typically set to 50 frames, which means reshaping $x_0$ to $(N//ws, ws, 128)$.
Finally, $x_0$ is projected onto \textit{latent reaction space}, shown in the middle of Fig.~\ref{fig:all}(b). 
This not only enhances the internal representation of each modality but also achieves semantic alignment between modalities:
\begin{equation}
\label{eq:Equation 3}
\begin{split}
f^{v} &= IntraCA(MLP(f^{VA},f^{AUs},f^{FE})), \\
f^{a} &= IntraCA(f^{MFCC},f^{AFR}), \\
x_0 &= MLP(Reshape(InterCA(f^{v},f^{a}))),
\end{split}
\end{equation}
where $IntraCA$ and $InterCA$ represent intra-
cross and inter-cross attention, respectively, $MLP$ is multi-layer perceptron and $x_0$ is the input of latent diffusion model.
The decoder, comprising Transformer encoder layers that designed to efficiently process parallel sequences while employing self-attention to ensure temporal consistency across the entire sequence, and two linear heads that reconstructs latent vectors into specific labels:
\begin{equation}
\label{eq:Equation 4}
\begin{split}
3dmm, emo &= MLP_1(encoder(x_0)), MLP_2(encoder(x_0)).
\end{split}
\end{equation}

We reshape $x_0$ to $(N, 128)$ after the diffusion process, and the Transformer models the contextual relationships between multiple windows in a clip to ensure temporal consistency in the generated reactions.
These labels include the 3DMM coefficients, denoted as $3dmm\in{\mathbb{R}}^{N\times58}$, and emotion, denoted as ${emo}\in{\mathbb{R}}^{N\times25}$. 
The basic idea behind 3DMM is to represent a 3D face as a linear combination of basic shapes and textures learned from a dataset of 3D scans.
To visualize the generated facial reactions, FaceVerseV2~\citep{wang2022faceverse} and PIRender~\citep{ren2021pirenderer} are utilized, enabling the conversion of 3DMM coefficients into 2D facial images that are conditioned on the reference portrait identity.
The emotion includes three attributes: VA is a 2-dimensional vector ranging from -1 to 1, AU is a 15-dimensional binary vector of 0 or 1, and FE is an 8-dimensional vector ranging from 0 to 1.

The final objective is as follows:\par
\begin{footnotesize} 
\begin{equation}
\label{eq:Equation 5}
\begin{split}
L_{mmt} &= w_1 * L_{mse} + w_2 * L_{l1},\\
L_{mse} &= \sum_{t=1}^{N}  \mathrm{MSE}\left ( 3dmm_{t}, \hat{3dmm}_{t}\right ), \\
L_{l1} &= \sum_{t=1}^{N}  \mathrm{L1}\left ( emo_{t},\hat{emo}_{t} \right ),\\
\end{split}
\end{equation}
\end{footnotesize}where $N$ is frame number of clip, $\mathrm{MSE}$ is Mean Squared Error loss function, $\mathrm{L1}$ is Mean Absolute Error loss function. 
$w_1$ and $w_2$ are both set to 10 in our implementation.
$3dmm_{t}$ and $\hat{3dmm}_{t}$ represent the true and predicted 3dmm coefficients of the listener, respectively, for the $t$-th frame. 
In the same way, $emo_{t}$ and $\hat{emo}_{t}$ represent the true and predicted $t$-th-frame emotion coefficients of the listener, respectively.

The MMT plays a crucial role in the ReactDiff framework. 
It processes the multi-modal inputs from the feature extraction module and provides structured latent representations for the subsequent conditional DDIM. 
This design enables the model to generate facial reactions that are both diverse and contextually appropriate.

\subsection{Latent Diffusion}
While MMT provides a solid foundation for processing multi-modal data, we further enhance our model's capabilities by incorporating a conditional diffusion process.
To achieve more realistic reaction generation, we add a conditional DDIM in Latent Reaction Space, between the encoder and decoder of MMT, shown in Fig.~\ref{fig:all}(c).

DDIM generates samples through an iterative denoising process, gradually refining the output over multiple steps. This iterative nature allows the model to explore a wider range of possible outputs and better cover the modes of data distribution. 
At each step of the reverse diffusion process, the model has the opportunity to correct errors made in previous steps, enabling it to generate more diverse and realistic samples.
Moreover, the conditioning mechanism enables our model to generate reactions that are not only diverse but also coherent with the speaker's audio-visual cues. 
By guiding the diffusion process with relevant context, we ensure that the generated reactions are appropriate and realistic responding to the speaker's behaviour.

Our DDIM takes as input a latent sequence vector $\mathbf{z} = (\mathbf{z}_{bc}, \mathbf{x}_T)$, which consists of the speaker's behaviour constraint vector $\mathbf{z}_{bc}$ and a noise map $\mathbf{x}_T$ by diffusion process.
Among them, $\mathbf{z}_{bc}\in{\mathbb{R}}^{(N//ws) \times1024}$ is obtained by applying MLP to the speaker's latent vector sequence (see Equation ~\ref{eq:Equation 6}), as strong constraints at the frame level will reduce the diversity of reactions, :
\begin{equation}
\label{eq:Equation 6}
\begin{split}
x_{0}^{'} &= \mathrm{Flatten}\left ( x_0 \right )   \in \mathbb{R}^{\left ( N//ws \right ) \times 128}, \\
\mathbf{z}_{bc} &= \mathrm{GroupNorm}\left ( \mathrm{MLP}\left ( x_{0}^{'} \right )   \right )   \in \mathbb{R}^{\left ( N//ws \right ) \times 1024}.
\end{split}
\end{equation}\par

To encode noise map $\mathbf{x}_T$, we run the deterministic forward process of DDIM using noise prediction network $\epsilon_{\theta}$ that is conditioned on $\mathbf{z}_{bc}$.
Specifically, the latent sequence vector $\mathbf{z}=(\mathbf{z}_{bc}, \mathbf{x}_T)$ can be reconstructed to the original frame clip by running the generative reverse process of DDIM in a deterministic manner:
\begin{equation}
\label{eq:Equation 7}
\begin{split}
p_{\theta}\left ( \mathbf{x}_{0:T} | \mathbf{z}_{bc} \right )= p\left ( \mathbf{x}_{T} \right )\prod_{t=1}^{T} p_{\theta}\left (\mathbf{x}_{t-1}|\mathbf{x}_t, \mathbf{z}_{bc}\right ).
\end{split}
\end{equation}
\par
This noise prediction network $\epsilon_{\theta}$ is a modified version of UNet following Preechakul et al.~\citep{preechakul2022diffusion}.
Training is done by optimizing $L_{eps}$ with respect to $\theta$:\par
\begin{equation}
\label{eq:Equation 8}
\begin{split}
L_{eps} = \sum_{t=1}^{T} \mathbb{E}_{\mathbf{x}_0, \epsilon_{t}} \left \| \epsilon _{\theta}\left ( \mathbf{x}_t, t, \mathbf{z}_{bc} \right)-\epsilon_{t} \right \| _1,
\end{split}
\end{equation}
where $\epsilon_{t} \in \mathbb{R}^{ws\times 128}\sim \mathcal{N}\left ( 0, \mathbf{I} \right )$, $\mathbf{x}_t=\sqrt{\alpha_t}\mathbf{x}_0 + \sqrt{1 -\alpha_t}\epsilon _t$ and $T$ is set to 20. 
For training, the noise map $\mathbf{x}_T$ is not needed, and MMT is frozen. 

The decoder part of our model is jointly composed of conditional Denoising Diffusion Implicit Models (DDIM) and Transformer Layers, ensuring a balance between diversity and appropriateness while leveraging their respective reconstruction capabilities. 

Inspired by PerFRDiff’s weight-editing paradigm, we further enhance ReactDiff by introducing Listener Reaction Shifts (LRS) to model personalized cognitive styles. While PerFRDiff employs a large DiT backbone to achieve high diversity, our method prioritizes appropriateness and synchronization through latent space standardization and multi-modal fusion. As shown in Table \ref{test_app}, \ref{test_div} and \ref{test_rea_syn}, ReactDiff (LRS) achieves the best FRDist (79.63), FRDvs (0.1380) and FRSyn (39.40), surpassing PerFRDiff by 16\%, 57\% and 13\%, respectively. This highlights the advantages of our lightweight yet effective architecture in real-time interaction scenarios.

Overall, our ReactDiff integrates feature extractor, MMT, and DDIM. 
The feature extractor assists the model in focusing on learning mappings of reactions.
MMT captures correlations between multi-modal inputs and aligns them semantically across spatiotemporal dimensions, focusing on the appropriateness of the generated facial reactions.
DDIM helps generate diverse and realistic facial reactions that are contextually appropriate by introducing noise to explore various possibilities and iteratively refining the output based on the speaker's audio-visual cues.
This approach ensures that the generated listener facial reactions are both varied and contextually appropriate, enhancing the realism, appropriateness, and diversity of the synthesized reactions.

\section{Experiments}
\subsection{Datasets}
We have used two datasets with video-conference setting provided by REACT2024 challenge~\citep{song2024react}.
\begin{itemize}
    \item \textbf{NoXI.} The Novice eXpert Interaction (NoXI) dataset~\citep{cafaro2017noxi} is a dyadic interaction dataset that is annotated during an information retrieval task targeting multiple languages, multiple topics, and the occurrence of unexpected situations. NoXI is a corpus of screen-mediated face-to-face interactions recorded at three locations (France, Germany and UK), spoken in seven languages (English, French, German, Spanish, Indonesian, Arabic and Italian) discussing a wide range of topics.
    \item \textbf{RECOLA.} The REmote COLlaborative and Affective (RECOLA)~\citep{ringeval2013introducing} dataset consists of 9.5 hours of audio, visual, and physiological recordings of online dyadic interactions between 46 French speaking participants.
\end{itemize}

\subsection{Implementation Detail}
Initially, we conduct feature extraction on both video and audio data. Our training strategy is divided into two stages. 
\begin{itemize}
    \item In the first stage, the processed sequences are synchronized along the temporal dimension and semantically integrated within MMT, ensuring a comprehensive interaction between the two modalities.
    \item In the second stage, we freeze the parameters of MMT and incorporate DDIM for joint training, utilizing speaker sequences as conditioning inputs.
\end{itemize}

We have trained MMT and ReactDiff on an NVIDIA A100 GPU with a batch size of 16 and learning rate of 5e-3, and 100 epochs both, respectively. Video frame resolution is $224\times 224$ and training time of two stages is 16h.
We use the AdamW optimizer~\citep{loshchilov2016sgdr} with $\beta_{1}= 0.9, \beta_{2}= 0.999$ and apply a weight decay of $5e-4$.
We generate $\alpha$ = 10 reaction sequences per speaker sequence for evaluation, evaluation metrics shown in the Appendix. 

\begin{table}[]
\centering
\caption{Appropriateness results of online and offline facial reaction generation task on the test set.}
\label{test_app}
\resizebox{\linewidth}{!}{
\renewcommand{\arraystretch}{1.5}
\begin{tabular}{c c c}
\bottomrule
{Method} & FRCorr($\uparrow$) & FRDist($\downarrow$) \\
\hline
GT & 8.73 & 0.00 \\
\hline
B\_Random & 0.05 & 237.21  \\

B\_Mime & 0.38 & 92.94  \\

B\_MeanSeq & 0.01 & 97.13 \\

B\_MeanFr & 0.00 & 97.86 \\
\hline
\multicolumn{3}{l}{Offline Results} \\
\hline
Trans-VAE~\citep{song2023react2023} & 0.03 & 92.81 \\
BeLFusion (k=1)~\citep{barquero2023belfusion} & 0.10 & 92.32 \\
BeLFusion (k=10)~\citep{barquero2023belfusion} & 0.12 & 91.60 \\

REGNN~\citep{xu2023reversible} & 0.19 & 84.54 \\

VQ-Diff~\citep{nguyen2024vector}   &0.30    &91.65 \\

Unifarn~\citep{liang2023unifarn}   &0.19    &98.51\\

Beamer~\citep{hoque2023beamer}    &0.11    &97.33\\

FRDiff~\citep{yu2023leveraging}    &0.14     &91.05 \\

PerFRDiff~\citep{zhu2024perfrdiff}   &\textbf{0.38}   &94.72 \\

Liu et.al~\citep{liu2024one} &0.22 &100.43 \\

ReactDiff (LRS)   &0.29     &\textbf{79.63} \\

ReactDiff & 0.26 & 86.70 \\
\hline
\multicolumn{3}{l}{Online Results} \\
\hline
Trans-VAE & 0.07 & 90.31\\

BeLFusion (k=1) & 0.12 & 91.11 \\

BeLFusion(k=10) & 0.12 & 91.45 \\

VQ-Diff    &\textbf{0.30}  &91.86\\

Liu et.al &0.22 &\textbf{88.32}  \\

ReactDiff & 0.24 & 89.17 \\
\bottomrule
\end{tabular}
}
\end{table}

\subsection{Comprehensive Evaluation of ReactDiff's Performance}

\begin{figure}[t]
    \centering
    \includegraphics[width=1\linewidth]{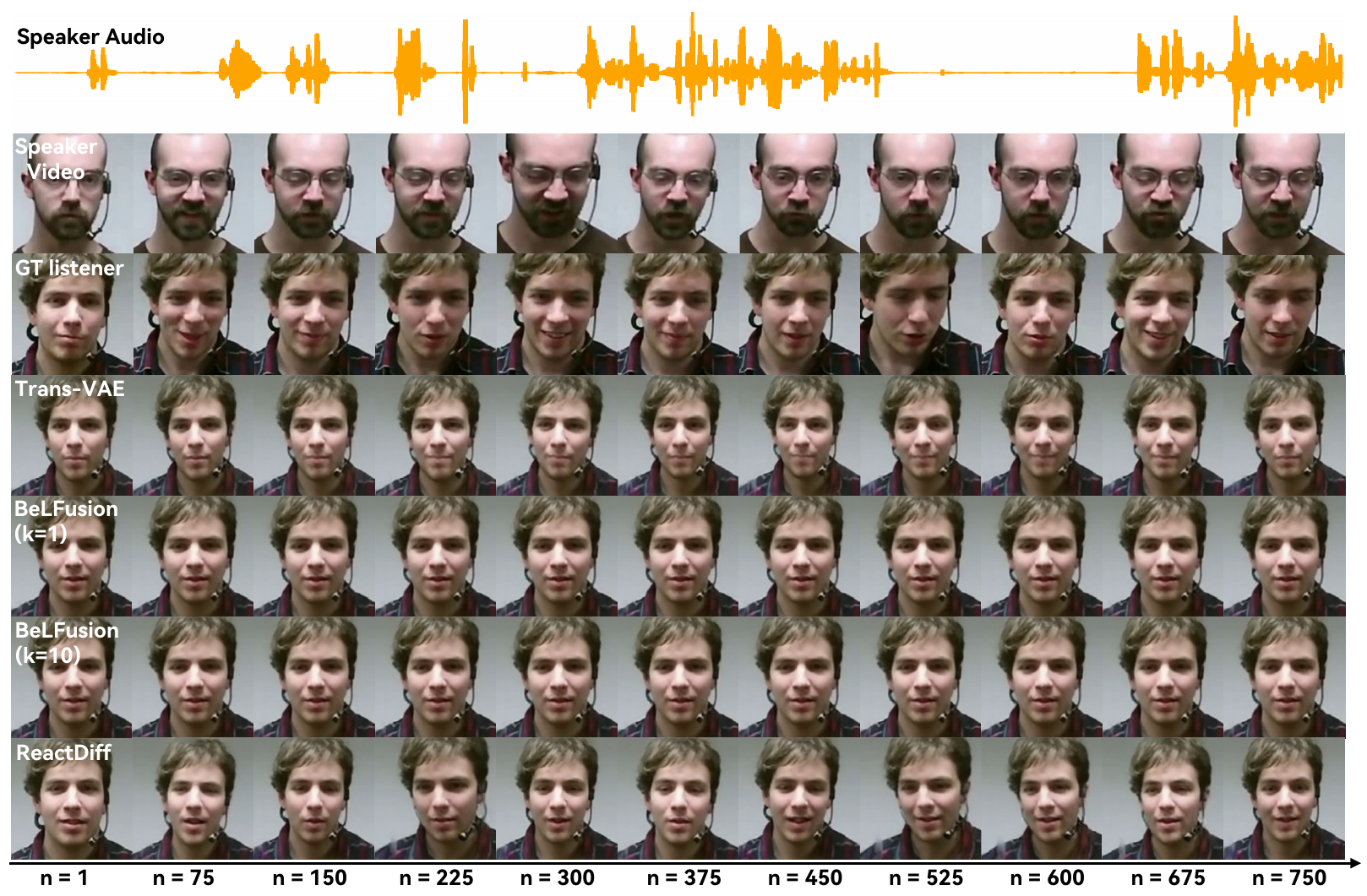}
    \caption{Qualitative comparison of generated listener reactions by models on the testing set and $n$ denotes frame numbers.}
    \label{fig:visual_contra}
\end{figure}

To demonstrate the performance of the ReactDiff framework, we conducted experiments on two sub-challenges of REACT2024~\citep{song2024react}: Offline and Online Multiple Appropriate Facial Reaction Generation (Offline and Online MAFRG).
Offline MAFRG task focuses on generating multiple appropriate facial reaction videos from the input speaker behaviour. Online MAFRG task focuses on the continuous generation of facial reaction frames based on current and previous speaker behaviour. Some reactions generated by offline ReactDiff are shown in Fig.~\ref{fig:visual_multi}.
The challenge establishes a set of naive baselines, namely B\_Random, B\_Mime, and B\_MeanSeq/B\_MeanFr. Specifically, B\_Random randomly samples $\alpha$ = 10 facial reaction sequences from a Gaussian distribution. B\_Mime generates facial reactions by mimicking the corresponding speaker’s facial expressions. For B\_MeanSeq and B\_MeanFr, the generated facial reactions are decided by the sequence-wise and frame-wise average reaction in the training set, respectively. Additionally, there are Trans-VAE~\citep{song2023react2023}, BeLFusion~\citep{barquero2023belfusion} and REGNN~\citep{xu2023reversible} baselines. The visualization results of ReactDiff and the above methods are shown in Fig.~\ref{fig:visual_contra}.

\subsubsection{Appropriateness result}

\begin{figure}[t]
    \centering
    \includegraphics[width=1\linewidth]{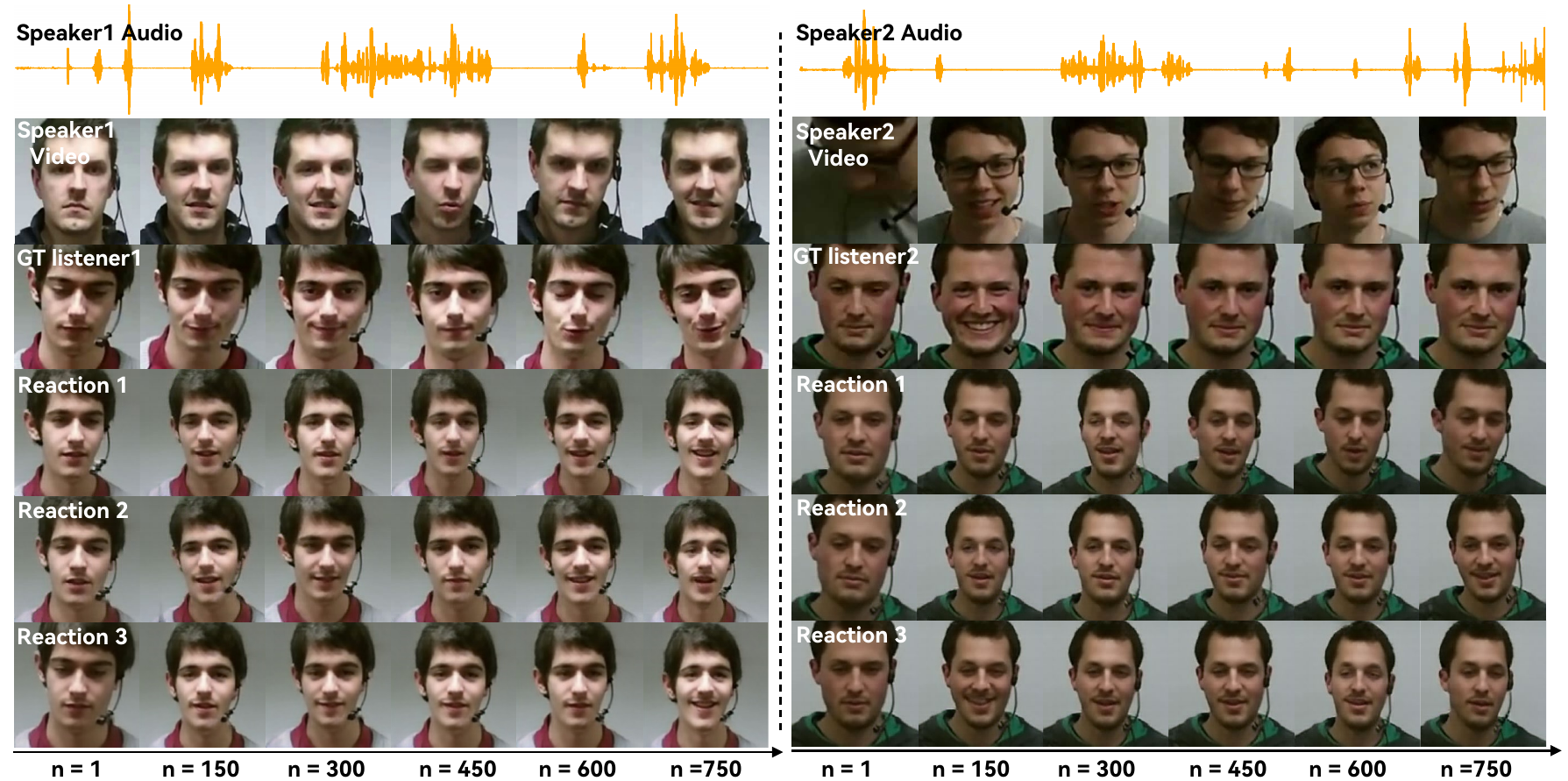}
    \caption{Two examples of generated listener reactions by ReactDiff on the test set and $n$ denotes frame numbers.}
    \label{fig:visual_multi}
\end{figure}

Table \ref{test_app} presents the appropriateness results for both online and offline facial reaction generation tasks on the test set, as measured by FRCorr (Facial Reaction Correlation) and FRDist (Facial Reaction Distance) metrics.
For the offline task, ReactDiff achieves the highest FRCorr score of 0.26 among all methods, significantly outperforming other approaches. This indicates that ReactDiff generates facial reactions are most correlated with ground truth reactions. In terms of FRDist, ReactDiff scores 86.70, which is competitive but slightly higher than REGNN (84.54).
In the online setting, ReactDiff maintains its superior performance with the highest FRCorr of 0.24, substantially outperforming other methods. The FRDist for ReactDiff in the online setting is 89.17, which is competitive with other approaches.
Compared to naive baselines (B\_Random, B\_Mime, B\_MeanSeq, B\_MeanFr), ReactDiff shows dramatic improvements in both metrics, demonstrating the effectiveness of its sophisticated architecture. It also significantly outperforms more advanced baselines like Trans-VAE and BeLFusion in terms of appropriateness.
It's worth noting that while the ground truth naturally achieves perfect scores (FRCorr of 8.73 and FRDist of 0.00), ReactDiff's performance represents the current state-of-the-art among generative models for this task.
As shown in Fig.~\ref{fig:visual_multi}, ReactDiff is particularly effective at generating appropriate facial reactions correlated with speaker reactions.

\subsubsection{Diversity result}
Table \ref{test_div} presents the diversity results for both online and offline facial reaction generation tasks on the test set, measured by three metrics: FRDiv (Diverseness among generated facial reactions), FRVar (Facial reaction variance), and FRDvs (Diversity among facial reactions generated from different speaker behaviors).
In the offline and online tasks, ReactDiff demonstrates superior performance across all diversity metrics.
Notably, while the ground truth shows zero diversity in FRDiv (as expected for a single ground truth), it does have non-zero FRVar and FRDvs, indicating natural variations in real facial reactions.
While B\_Random achieves high diversity scores, this is expected as random sampling naturally produces diverse outputs. However, these random reactions lack appropriateness and realism, which are crucial for practical applications. This highlights that diversity alone is not sufficient, the generated reactions must also be contextually appropriate and realistic.
As shown in Fig.~\ref{fig:visual_multi}, ReactDiff can generate a wide range of diverse facial reactions, both for individual speakers and across different speaker behaviors. 
This diversity is crucial for creating realistic and varied interactions in dyadic scenarios.

\begin{table}[htbp]
\centering
\caption{Diversity results of online and offline facial reaction generation task on the test set.}
\label{test_div}
\resizebox{\linewidth}{!}{
\renewcommand{\arraystretch}{1.5}
\begin{tabular}{c c c c}
\bottomrule
Method & FRDiv($\uparrow$) & FRVar($\uparrow$) & FRDvs($\uparrow$) \\

\hline
GT & 0.0000 & 0.0724 & 0.2483 \\
\hline
B\_Random & 0.1667 & 0.0833 & 0.1667 \\

B\_Mime & 0.0000 & 0.0724 & 0.2483 \\

B\_MeanSeq & 0.0000 & 0.0000 & 0.0000 \\

B\_MeanFr & 0.0000 & 0.0000 & 0.0000\\
\hline
\multicolumn{4}{l}{Offline Results} \\
\hline
Trans-VAE~\citep{song2023react2023} & 0.0008 & 0.0002 & 0.0006 \\
BeLFusion (k=1)~\citep{barquero2023belfusion} & 0.0068 & 0.0073 & 0.0094 \\
BeLFusion (k=10)~\citep{barquero2023belfusion} & 0.0105 & 0.0082 & 0.0116 \\

REGNN~\citep{xu2023reversible} & 0.0007 & 0.0061 & 0.0342 \\

VQ-Diff~\citep{nguyen2024vector}    &0.0743    &0.0348    &0.0745 \\

Unifarn~\citep{liang2023unifarn}    &0.0819    &0.0760     &0.0259 \\

Beamer~\citep{hoque2023beamer}    &0.0508    &0.0374      &0.0196 \\

FRDiff~\citep{yu2023leveraging}     &0.0690      &0.0443   &0.0108 \\

PerFRDiff~\citep{zhu2024perfrdiff}      &0.1368     &\textbf{0.2191}   &0.0879 \\

Liu et.al~\citep{liu2024one} &\textbf{0.1675} &0.0535 &\textbf{0.1385} \\

ReactDiff (LRS)  &0.0334     &0.0401     &0.1380 \\

ReactDiff & 0.0940 & 0.0462 & 0.1012 \\
\hline
\multicolumn{4}{l}{Online Results} \\
\hline
Trans-VAE & 0.0064 & 0.0012 & 0.0009  \\

BeLFusion (k=1) & 0.0083 & 0.0079 & 0.0103  \\

BeLFusion(k=10) & 0.0112 & 0.0082 & 0.0120  \\

VQ-Diff  &0.0737 &0.0346 &0.0743 \\
Liu et.al  &\textbf{0.1029} &0.0387 &\textbf{0.1065} \\

ReactDiff  & 0.0900 & \textbf{0.0452} & 0.1005 \\
\bottomrule
\end{tabular}
}
\end{table}

\subsubsection{Realism and synchrony result}
Table \ref{test_rea_syn} presents the results for realism and synchrony in both online and offline facial reaction generation tasks on the test set. These aspects are evaluated using two metrics: FRRea (Facial Reaction Realism) and FRSyn (Synchrony between generated facial reactions and speaker behaviors).
In the offline setting, ReactDiff achieves an FRRea score of 66.72, which is the best among the compared methods, outperforming Trans-VAE (67.74) and BeLFusion variants (79.60 and 79.59). 
For the online task, ReactDiff's FRRea score increases to 73.20, which is higher than its offline performance but not as good as Trans-VAE (69.19).
This suggests that maintaining high realism in real-time generation is more challenging. In terms of synchrony, ReactDiff demonstrates consistent performance across both offline (45.40) and online (45.26) settings, though there's room for improvement compared to the best-performing methods like REGNN (41.35). These results indicate that while ReactDiff maintains competitive performance in terms of realism and synchrony, there are opportunities for enhancement, particularly in online scenarios and in balancing the trade-off between realism and synchrony.
Notably, B\_Mime achieves better synchrony scores because it directly copies speaker expressions, ensuring tight temporal alignment. 
However, this approach ignores the natural variation in listener reactions and fails to capture genuine interactive dynamics.
Similarly, B\_Random may occasionally produce realistic-looking expressions by chance, but these are not contextually appropriate. ReactDiff makes a conscious trade-off, prioritizing appropriate and diverse reactions while maintaining acceptable levels of realism and synchrony.

\begin{table}[]
\centering
\caption{Realism and synchrony results of online and offline facial reaction generation task on the test set.}
\label{test_rea_syn}
\resizebox{\linewidth}{!}{
\renewcommand{\arraystretch}{1.5}
\begin{tabular}{c c c}
\bottomrule
Method & FRRea($\downarrow$) & FRSyn($\downarrow$) \\

\hline
GT & 53.96 & 47.69 \\
\hline
B\_Random & - & 43.84 \\

B\_Mime & - & 38.54 \\

B\_MeanSeq & - & 45.28 \\

B\_MeanFr & - & 49.00 \\
\hline
\multicolumn{3}{l}{Offline Results} \\
\hline
Trans-VAE~\citep{song2023react2023} & 67.74 & 43.75 \\
BeLFusion (k=1)~\citep{barquero2023belfusion} & 79.60 & 44.94 \\
BeLFusion (k=10)~\citep{barquero2023belfusion} & 79.59 & 44.87 \\

REGNN~\citep{xu2023reversible} & - & 41.35 \\

VQ-Diff~\citep{nguyen2024vector}      &- &45.33\\

Unifarn~\citep{liang2023unifarn}        &-       &46.11 \\

Beamer~\citep{hoque2023beamer}       &-      &48.12 \\

FRDiff~\citep{yu2023leveraging}       &69.37      &47.66 \\

PerFRDiff~\citep{zhu2024perfrdiff}     & \textbf{47.62}     &45.28\\

Liu et.al~\citep{liu2024one} &- & 44.54 \\

ReactDiff (LRS)  & 76.89    &\textbf{39.40} \\

ReactDiff & 66.72 & 45.40 \\
\hline
\multicolumn{3}{l}{Online Results} \\
\hline
Trans-VAE & \textbf{69.19} & 44.65 \\

BeLFusion (k=1) & - & 45.17 \\

BeLFusion(k=10) & - & 44.89 \\

VQ-Diff   &-   &45.18 \\
Liu et.al &- &\textbf{44.41} \\

ReactDiff & 73.20 & 45.26 \\
\bottomrule
\end{tabular}
}
\end{table}
Overall, these results indicate that ReactDiff maintains competitive performance in terms of realism and synchrony. While it may not always achieve the best scores in these metrics, it demonstrates a good balance, especially considering its superior performance in appropriateness and diversity as shown in previous tables.

\subsection{Ablation Study}
Table \ref{tab:ablation} illustrates the impact of various components in the ReactDiff framework. Ablation of the Multi-Modality Transformer (MMT) significantly reduces appropriateness and diversity, while maintaining realism. 
Conversely, removing the conditional DDIM which operates in the latent reaction space between MMT's encoder and decoder, dramatically increases appropriateness but eliminates diversity.
Exclusion of video input decreases appropriateness and diversity, with a slight reduction in realism.
Similarly, omitting audio input marginally affects appropriateness and diversity, with minimal impact on realism.
The complete ReactDiff framework exhibits balanced performance across metrics, achieving the second-highest appropriateness score, optimal diversity scores, and commendable realism. These results underscore the crucial role of each component in the model's performance. 
The MMT is vital for balancing appropriateness and synchrony, while the Diffusion component enhances realism and diversity. Multi-modal inputs (video and audio) are essential for generating appropriate and diverse reactions. In summary, the full ReactDiff framework demonstrates the most effective balance across evaluation metrics, highlighting the synergistic effect of its components in generating appropriate, diverse, and realistic facial reactions.

\begin{table*}[]
\centering
\caption{Ablation results of offline facial reaction generation task on the test set.}
\label{tab:ablation}
\begin{tabular}{c c c c c c c c c c}
\bottomrule
\multirow{2}*{Method} &\multicolumn{2}{c}{Appropriateness} & \multicolumn{3}{c}{Diversity} & Realism & Synchrony \\
\cline{2-8}
 ~ & FRCorr($\uparrow$) & FRDist($\downarrow$) & FRDiv($\uparrow$) & FRVar($\uparrow$) & FRDvs($\uparrow$) & FRRea($\downarrow$) & FRSyn($\downarrow$) \\
\hline
ReactDiff & 0.26 & 86.70 & 0.0940 & 0.0462 & 0.1012 & 66.72 & 45.40 \\

(w/o) MMT & 0.02 & 162.38 & 0.0014 & 0.0008 & 0.0015 & 72.21 & 43.75 \\

(w/o) Diffusion & 0.84 & 80.10 & 0.0000 & 0.0247 & 0.1349 & 78.00 & 45.02 \\

(w/o) Video & 0.13 & 92.09 & 0.0292 & 0.0149 & 0.0294 & 82.12 & 45.18 \\

(w/o) Audio & 0.24 & 87.27 & 0.0806 & 0.0395 & 0.0866 & 78.26 & 45.43 \\

\bottomrule
\end{tabular}
\end{table*}

\section{DISCUSSION AND CONCLUSION}

In this work, have we proposed ReactDiff, a framework for generating appropriate, diverse, and realistic facial reactions in dyadic interactions. ReactDiff addresses key challenges by leveraging a Multi-Modality Transformer for cross-modal representations and a conditional diffusion model in the latent space. Experiments on the REACT2024 challenge datasets demonstrate ReactDiff's superior performance compared to state-of-the-art baselines, validating the effectiveness of its architecture.
ReactDiff shows promise for various applications, including virtual assistants in healthcare consultations, where appropriate emotional responses are crucial for patient comfort; educational AI tutors that can provide engaging feedback; and customer service avatars capable of natural emotional interactions. The framework's ability to generate diverse yet appropriate reactions makes it particularly suitable for long-term Human-AI interactions where repetitive behaviors would diminish engagement.

While promising, ReactDiff has limitations to address in future work. These include pose jitter in 3DMM reconstruction, suboptimal compatibility of the FaceVerse model with non-Asian faces, ethnicity or age-related biases in training data and the need for real-time optimization. Addressing these issues could further advance facial reaction generation, bringing us closer to natural and engaging human-computer interactions.
The Multi-Modality Transformer efficiently learns cross-modal representations, while the conditional diffusion model enables diverse and realistic reactions guided by the speaker's context. This combination proves effective in balancing appropriateness, diversity, and realism in generated facial reactions.

\appendix
\section{Evaluation Metrics}
\label{appendix}

\begin{itemize}
\item Facial reaction correlation (FRCorr).
we compute the correlation between each generated facial reaction and its most similar appropriate real facial reaction, and final correlation score for evaluation is obtained by averaging as:
\begin{equation}
\label{eq:Equation 9}
\begin{split}
\mathrm{FRCorr} = \frac{\sum_{n=1}^{N} \sum_{i=1}^{\alpha } \mathrm{Max}\left ( \mathrm{CCC}\left ( y_{i}^{n}, \hat{y}^{n} \right )  \right )   }{N} 
\end{split}
\end{equation}where $N$ is frame number of clip, $CCC$ denotes the Concordance Correlation Coefficient~\citep{lawrence1989concordance}, $y_{i}^{n}$ denotes the $n$-th frame of the $i$-th generated listener facial reaction ($emotion$ in paper, includes multi-channel facial attributes), $\hat{y}^{n}$ denotes the $n$-th frame of a set of appropriate real listener facial reactions corresponding to a speaker behaviour (The similarity matrix is computed for each speaker reaction sequence, and the listener reaction corresponding to multiple speakers whose similarity with the $i$-th speaker exceeds the threshold is utilized as a set of appropriate facial listener reactions corresponding to the $i$-th speaker).

\item Facial Reaction Distance (FRDist). 
we compute the Dynamic Time Warping (DTW)~\citep{berndt1994using} distance between the generated facial reaction and its most similar appropriate real facial reaction, and final distance score for evaluation is obtained by averaging as:
\begin{equation}
\label{eq:Equation 10}
\begin{split}
\mathrm{FRDist} = \frac{\sum_{n=1}^{N} \sum_{i=1}^{\alpha } \mathrm{Min}\left ( \mathrm{DTW}\left ( y_{i}^{n}, \hat{y}^{n} \right )  \right )   }{N} 
\end{split}
\end{equation}

\item Diverseness among generated facial reactions (FRDiv).
we evaluate model's diverseness ability by calculating the sum of the MSE among every pair of the generated facial reactions corresponding to each input speaker behaviour as:
\begin{equation}
\label{eq:Equation 11}
\begin{split}
\mathrm{FRDiv} = \frac{\sum_{n=1}^{N} \sum_{i=1}^{\alpha -1}  \sum_{j=i+1}^{\alpha } \left ( y_{i}^{n}- y_{j}^{n} \right )^2   }{N} 
\end{split}
\end{equation}

\item Facial reaction variance (FRVar).
We evaluate the variance of each generated facial reaction, which is obtained by computing the variation across all of its frames, and final facial reaction diversity is obtained by averaging the variance values of all generated facial reactions:
\begin{equation}
\label{eq:Equation 12}
\begin{split}
\mathrm{FRVar} = \frac{\sum_{n=1}^{N} \sum_{i=1}^{\alpha } \mathrm{var} \left ( y_{i}^{n} \right )   }{N\times \alpha } 
\end{split}
\end{equation}

\item Diversity among facial reactions generated from different speaker behaviours (FRDvs).
we finally evaluate the diversity of the generated facial reactions corresponding to different speaker behaviours as:
\begin{equation}
\label{eq:Equation 13}
\begin{split}
\mathrm{FRDvs} = \frac{\sum_{i=1}^{\alpha } \sum_{n=1}^{N-1}  \sum_{m=n+1}^{N} \mathrm{MSE}  \left ( y_{i}^{n}, y_{j}^{m} \right )   }{\alpha  N\left ( N- 1 \right ) }  
\end{split}
\end{equation}

\item Facial reaction realism (FRRea).
The realism score is obtained by computing Frechet Inception Distance (FID)~\citep{heusel2017gans} between the distribution Dis($Y$) of the generated facial reactions and the distribution Dis($\hat{Y}$) of the corresponding appropriate real facial reactions as:
\begin{equation}
\label{eq:Equation 14}
\begin{split}
\mathrm{FRRea} = \mathrm{FID} \left ( \mathrm{Dis}\left ( Y \right ) , \mathrm{Dis}\left ( \hat{Y}  \right )    \right )  
\end{split}
\end{equation}where $Y$ denotes a sequence of generated facial reactions, $\hat{Y}$ denotes a sequence of real listener facial reactions corresponding to speaker behaviour.

\item Synchrony between generated facial reactions and speaker behaviours (FRSyn).
we compute the Time Lagged Cross Correlation (TLCC)~\citep{shen2015analysis} to evaluate the synchrony between the input speaker behaviour and the corresponding generated facial reaction.
\begin{equation}
\label{eq:Equation 15}
\begin{split}
\mathrm{FRSyn} = \frac{\sum_{n=1}^{N} \sum_{i=1}^{\alpha } \left ( y_{i}^{n}, y_{S}^{n} \right )   }{N}  
\end{split}
\end{equation}where $y_{S}^{n}$ denotes $n$-th frame of the speaker reaction.
\end{itemize}

\printcredits

\bibliographystyle{cas-model2-names}

\bibliography{cas-refs}



\end{document}